\documentclass[11pt,a4paper]{article}
\usepackage[hyperref]{naaclhlt2019}
\usepackage{times}
\usepackage{latexsym}

\usepackage{url}

\usepackage{times}
\usepackage{latexsym}
\usepackage{amsmath,amssymb}
\usepackage{mathtools}
\usepackage{csquotes}
\usepackage{graphicx}
\usepackage{multirow}
\usepackage{listings}
\usepackage{comment}
\usepackage[normalem]{ulem}
\usepackage{enumitem}
\usepackage{bbm}
\usepackage{wrapfig}

\usepackage{setspace}

\definecolor{Code}{rgb}{0,0,0}
\definecolor{Decorators}{rgb}{0.5,0.5,0.5}
\definecolor{Numbers}{rgb}{0.5,0,0}
\definecolor{MatchingBrackets}{rgb}{0.25,0.5,0.5}
\definecolor{Keywords}{rgb}{0,0,1}
\definecolor{self}{rgb}{0,0,0}
\definecolor{Strings}{rgb}{0,0.63,0}
\definecolor{Comments}{rgb}{0,0.63,1}
\definecolor{Backquotes}{rgb}{0,0,0}
\definecolor{Classname}{rgb}{0,0,0}
\definecolor{FunctionName}{rgb}{0,0,0}
\definecolor{Operators}{rgb}{0,0,0}
\definecolor{Background}{rgb}{0.98,0.98,0.98}
\lstdefinelanguage{Python}{numbers=left,numberstyle=\footnotesize,numbersep=1em,
xleftmargin=1em,framextopmargin=2em,framexbottommargin=2em,showspaces=false,
showtabs=false,showstringspaces=false,frame=l,
tabsize=4,
basicstyle=\ttfamily\small\setstretch{1},backgroundcolor=\color{Background},
commentstyle=\color{Comments}\slshape,
stringstyle=\color{Strings},morecomment=[s][\color{Strings}]{"""}{"""},morecomment=[s][\color{Strings}]{'''}{'''},morecomment=[l][\color{Strings}]{\#},
morekeywords={import,from,class,def,for,while,if,is,in,elif,else,not,and,or,print,break,continue,return,True,False,None,access,as,,del,except,exec,finally,global,import,lambda,pass,print,raise,try,assert},keywordstyle={\color{Keywords}\bfseries},
morekeywords={[2]@invariant,pylab,numpy,np,scipy},keywordstyle={[2]\color{Decorators}\slshape},emph={self},emphstyle={\color{self}\slshape},
}

\newcommand\eg{\textit{e.g.}}
\newcommand\ie{\textit{i.e.}}
\newcommand\fren{\texttt{fr-en}}
\newcommand\deen{\texttt{de-en}}
\newcommand\csen{\texttt{cs-en}}
\newcommand\unk{\texttt{<unk>}}

\usepackage[acronym,shortcuts,acronymlists={hidden}]{glossaries}

\newacronym{nlp}{NLP}{natural language processing}
\newacronym{nmt}{NMT}{neural machine translation}
\newacronym{smt}{SMT}{Statistical Machine Translation}
\newacronym{mt}{MT}{machine translation}
\newacronym{lm}{LM}{Language Modeling}
\newacronym{bpe}{BPE}{Byte-Pair Encoding}
\newacronym{dataset}{MTNT}{Machine Translation of Noisy Text}
\newacronym{mtnt}{MTNT}{Machine Translation of Noisy Text}
\newacronym{oov}{OOV}{out-of-vocabulary}
\newacronym{nll}{NLL}{negative log likelihood}
\newacronym{rdb}{RDchrF}{relative decrease in chrF}
\newacronym{seq2seq}{seq2seq}{sequence-to-sequence}

\DeclareMathOperator*{\argmax}{arg\,max}
\renewcommand\cite{\citep}
\newcommand\codeurl{\url{https://github.com/pmichel31415/translate/tree/paul/pytorch_translate/research/adversarial/experiments}}
\newcommand\teapoturl{\url{https://github.com/pmichel31415/teapot-nlp}}
\newcommand\goodscore{chrF}
\newcommand\unconstrained{Unconstrained}
\newcommand\knn{kNN}
\newcommand\unkonly{CharSwap}

\DeclareMathOperator*{\w}{\mathbf{w}}
\DeclareMathOperator*{\Ladv}{\mathcal L_{\text{adv}}}
\newcommand{\bigO}[1]{\mathcal{O}(#1)}
\DeclareMathOperator*{\stgt}{s_{\text{tgt}}}
\DeclareMathOperator*{\ssrc}{s_{\text{src}}}
\DeclareMathOperator*{\dtgt}{d_{\text{tgt}}}

\title{On Evaluation of Adversarial Perturbations \\ for Sequence-to-Sequence Models}

\aclfinalcopy 

\author{
  Paul Michel$^1$, Xian Li$^2$, Graham Neubig$^1$, Juan Miguel Pino$^2$  \\
$^1$Language Technologies Institute, Carnegie Mellon University\\
$^2$Facebook AI\\
  {\tt \{pmichel1,gneubig\}@cs.cmu.edu},\\{\tt \{xianl,juancarabina\}@fb.com} \\ }

\begin{document}
\maketitle
\begin{abstract}
Adversarial examples --- perturbations to the input of a model that elicit large changes in the output --- have been shown to be an effective way of assessing the robustness of \ac{seq2seq} models.
However, these perturbations only indicate weaknesses in the model if they do not change the input so significantly that it legitimately results in changes in the expected output.
This fact has largely been ignored in the evaluations of the growing body of related literature.
Using the example of untargeted attacks on \ac{mt}, we propose a new evaluation framework for adversarial attacks on \ac{seq2seq} models that takes the semantic equivalence of the pre- and post-perturbation input into account. Using this framework, we demonstrate that existing methods may not preserve meaning in general, breaking the aforementioned assumption that source side perturbations should not result in changes in the expected output.
We further use this framework to demonstrate that adding additional constraints on attacks allows for adversarial perturbations that are more meaning-preserving, but nonetheless largely change the output sequence.
Finally, we show that performing untargeted adversarial training with meaning-preserving attacks is beneficial to the model in terms of adversarial robustness, without hurting test performance.
\footnote{A toolkit implementing our evaluation framework is released at \teapoturl{}.}
\end{abstract}

\section{Introduction}

Attacking a machine learning model with adversarial perturbations is the process of making changes to its input to maximize an adversarial goal, such as mis-classification \cite{Szegedy2013IntriguingPO} or mis-translation \cite{zhao2018generating}.
These attacks provide insight into the vulnerabilities of machine learning models and their brittleness to samples outside the training distribution. Lack of robustness to these attacks poses security concerns to safety-critical applications, \eg{} self-driving cars \cite{bojarski2016end}.

Adversarial attacks were first defined and investigated for computer vision systems (\citet{Szegedy2013IntriguingPO,Goodfellow2014ExplainingAH,MoosaviDezfooli2016DeepFoolAS} inter alia), where the input space is continuous, making minuscule perturbations largely imperceptible to the human eye.
In discrete spaces such as natural language sentences, the situation is more problematic; even a flip of a single word or character is generally perceptible by a human reader.
Thus, most of the mathematical framework in previous work is not directly applicable to discrete text data.
Moreover, there is no canonical distance metric for textual data like the $\ell_p$ norm in real-valued vector spaces such as images, and evaluating the level of semantic similarity between two sentences is a field of research of its own  \cite{cer-EtAl:2017:SemEval}.
This elicits a natural question: \textit{what does the term ``adversarial perturbation'' mean in the context of \ac{nlp}}?

We propose a simple but natural criterion for adversarial examples in \ac{nlp}, particularly untargeted\footnote{Here we use the term untargeted in the same sense as \cite{Ebrahimi2018OnAE}: an attack whose goal is simply to decrease performance with respect to a reference translation.} attacks on \ac{seq2seq} models: \emph{adversarial examples should be meaning-preserving on the source side, but meaning-destroying on the target side}.
The focus on explicitly evaluating meaning preservation is in contrast to previous work on adversarial examples for \ac{seq2seq} models \cite{belinkov2018synthetic,zhao2018generating,cheng2018seq2sick,Ebrahimi2018OnAE}.
Nonetheless, this feature is extremely important; given two sentences with equivalent meaning, we would expect a good model to produce two outputs with equivalent meaning.
In other words, any meaning-preserving perturbation that results in the model output changing drastically highlights a fault of the model.

A first technical contribution of this paper is to lay out a method for formalizing this concept of meaning-preserving perturbations (\S\ref{sec:eval_adv_attacks}).
This makes it possible to evaluate the effectiveness of adversarial attacks or defenses either using gold-standard human evaluation, or approximations that can be calculated without human intervention.
We further propose a simple method of imbuing gradient-based word substitution attacks (\S\ref{sec:attack_paradigm}) with simple constraints aimed at increasing the chance that the meaning is preserved (\S\ref{sec:constraints}).

Our experiments are designed to answer several questions about meaning preservation in \ac{seq2seq} models.
First, we evaluate our proposed ``source-meaning-preserving, target-meaning-destroying'' criterion for adversarial examples using both manual and automatic evaluation (\S\ref{sec:corr_human_auto}) and find that a less widely used evaluation metric (\goodscore{}) provides significantly better correlation with human judgments than the more widely used BLEU and METEOR metrics.
We proceed to perform an evaluation of adversarial example generation techniques, finding that \goodscore{} does help to distinguish between perturbations that are more meaning-preserving across a variety of languages and models (\S\ref{sec:attack_results}).
Finally, we apply existing methods for adversarial training to the adversarial examples with these constraints and show that making adversarial inputs more semantically similar to the source is beneficial for robustness to adversarial attacks and does not decrease test performance on the original data distribution (\S\ref{sec:adv_train}).

\section{A Framework for Evaluating Adversarial Attacks}
\label{sec:eval_adv_attacks}

In this section, we present a simple procedure for evaluating adversarial attacks on \ac{seq2seq} models. We will use the following notation: $x$ and $y$ refer to the source and target sentence respectively. We denote $x$'s translation by model $M$ as $y_M$. Finally, $\hat x$ and $\hat y_M$ represent an adversarially perturbed version of $x$ and its translation by $M$, respectively. The nature of $M$ and the procedure for obtaining $\hat x$ from $x$ are irrelevant to the discussion below.

\subsection{The Adversarial Trade-off}
\label{sec:adv_tradeoff}

The goal of adversarial perturbations is to produce failure cases for the model $M$. Hence, the evaluation must include some measure of the \emph{target similarity} between $y$ and $y_{M}$, which we will denote $\stgt(y, \hat y_M)$.
However, if no distinction is being made between perturbations that preserve the meaning and those that don't, a sentence like ``he's very \textit{friendly}'' is considered a valid adversarial perturbation of ``he's very \textit{adversarial}'', even though its meaning is the opposite.
Hence, it is crucial, when evaluating adversarial attacks on \ac{mt} models, that the discrepancy between the original and adversarial input sentence be quantified in a way that is sensitive to meaning. Let us denote such a \emph{source similarity} score $\ssrc(x,\hat x)$.

Based on these functions, we define the \emph{target relative score decrease} as:

\begin{equation}
\dtgt(y, y_M, \hat y_M)=
\begin{cases}
    0 \text{ if } \small\stgt(y, \hat y_M) \ge \stgt(y, y_M) \\
    \frac{\stgt(y, y_M)-\stgt(y, \hat y_M)}{\stgt(y, y_M)} \text{ otherwise}
\end{cases}
\end{equation}

The choice to report the \emph{relative} decrease in $\stgt$ makes scores comparable across different models or languages\footnote{Note that we do not allow negative $\dtgt$ to keep all scores between 0 and 1.}. For instance, for languages that are comparatively easy to translate (\eg{} French-English), $\stgt$ will be higher in general, and so will the gap between $\stgt(y, y_M)$ and $\stgt(y, \hat{y}_M)$. However this does not necessarily mean that attacks on this language pair are more effective than attacks on a ``difficult'' language pair (\eg{} Czech-English) where $\stgt$ is usually smaller.

We recommend that both $\ssrc$ and $\dtgt$ be reported when presenting adversarial attack results. However, in some cases where a single number is needed, we suggest reporting the attack's \emph{success} $\mathcal S\coloneqq \ssrc + \dtgt $.
The interpretation is simple: $\mathcal S>1 \Leftrightarrow\dtgt>1-\ssrc$, which means that the attack has destroyed the target meaning ($\dtgt$) more than it has destroyed the source meaning ($1-\ssrc$).

Importantly, this framework can be extended beyond strictly meaning-preserving attacks. For example, for targeted keyword introduction attacks \cite{cheng2018seq2sick,Ebrahimi2018OnAE}, the same evaluation framework can be used if $\stgt$ (resp. $\ssrc$) is modified to account for the presence (resp. absence) of the keyword (or its translation in the source). Similarly this can be extended to other tasks by adapting $\stgt$  (\eg{} for classification one would use the zero-one loss, and adapt the success threshold).

\subsection{Similarity Metrics}
\label{sec:eval_metrics}

Throughout \S\ref{sec:adv_tradeoff}, we have not given an exact description of the semantic similarity scores $\ssrc$ and $\stgt$. Indeed, automatically evaluating the semantic similarity between two sentences is an open area of research and it makes sense to decouple the definition of adversarial examples from the specific method used to measure this similarity. In this section, we will discuss manual and automatic metrics that may be used to calculate it.

\subsubsection{Human Judgment}
\label{sec:human_judgement}

Judgment by speakers of the language of interest is the \textit{de facto} gold standard metric for semantic similarity. Specific criteria such as adequacy/fluency \cite{Ma2006CorpusSF}, acceptability \cite{Goto2013OverviewOT}, and 6-level semantic similarity \cite{cer-EtAl:2017:SemEval} have been used in evaluations of \ac{mt} and sentence embedding methods.
In the context of adversarial attacks, we propose the following 6-level evaluation scheme, which is motivated by previous measures, but designed to be (1) symmetric, like \citet{cer-EtAl:2017:SemEval}, (2) and largely considers meaning preservation but at the very low and high levels considers fluency of the output\footnote{This is important to rule out nonsensical sentences and distinguish between clean and ``noisy'' paraphrases (\eg{} typos, non-native speech\ldots). We did not give annotators additional instruction specific to typos.}, like \citet{Goto2013OverviewOT}:

{
\begin{center}
\framebox{
\begin{minipage}{0.9\columnwidth}
How would you rate the similarity between the meaning of these two sentences?
\begin{enumerate}[itemsep=-4pt]
\setcounter{enumi}{-1}
\item The meaning is completely different or one of the sentences is meaningless
\item The topic is the same but the meaning is different
\item Some key information is different
\item The key information is the same but the details differ
\item Meaning is essentially equal but some expressions are unnatural
\item Meaning is essentially equal and the two sentences are well-formed English\footnote{Or the language of interest.}
\end{enumerate}
\end{minipage}
}
\end{center}
}

\subsubsection{Automatic Metrics}
\label{sec:auto_metrics}

Unfortunately, human evaluation is expensive, slow and sometimes difficult to obtain, for example in the case of low-resource languages. This makes automatic metrics that do not require human intervention  appealing for experimental research.
This section describes 3 evaluation metrics commonly used as alternatives to human evaluation, in particular to evaluate translation models.%
\footnote{
Note that other metrics of similarity are certainly applicable within the overall framework of \S\ref{sec:human_judgement}, but we limit our examination in this paper to the three noted here.
}

\textbf{BLEU:} \cite{papineni-EtAl:2002:ACL} is an automatic metric based on n-gram precision coupled with a penalty for shorter sentences. It relies on exact word-level matches and therefore cannot detect synonyms or morphological variations.

\textbf{METEOR:} \cite{denkowski:lavie:meteor-wmt:2014} first estimates alignment between the two sentences and then computes unigram F-score (biased towards recall) weighted by a penalty for longer sentences. Importantly, METEOR uses stemming, synonymy and paraphrasing information to perform alignments. On the downside, it requires language specific resources.

\textbf{chrF:} \cite{popovic:2015:WMT} is based on the character $n$-gram F-score. In particular we will use the chrF2 score (based on the F2-score --- recall is given more importance), following the recommendations from \citet{popovic:2016:WMT}. By operating on a sub-word level, it can reflect the semantic similarity between different morphological inflections of one word (for instance), without requiring language-specific knowledge which makes it a good one-size-fits-all alternative.

Because multiple possible alternatives exist, it is important to know which is the best stand-in for human evaluation.
To elucidate this, we will compare these metrics to human judgment in terms of Pearson correlation coefficient on outputs resulting from a variety of attacks in \S\ref{sec:corr_human_auto}.

\section{Gradient-Based Adversarial Attacks}
\label{sec:attacks}

In this section, we overview the adversarial attacks we will be considering in the rest of this paper.

\subsection{Attack Paradigm}
\label{sec:attack_paradigm}

\begin{table*}[!h]
\centering
{
\begin{tabular}{ll}
\hline\hline
Original & {\bf Pourquoi} faire cela ? \\
English gloss & {\bf Why} do this? \\
\unconstrained{} & {\color{red}\bf construisant} {\color{orange}(English: building)} faire cela ? \\ 
\knn{} & {\color{red}\bf interrogez} {\color{orange}(English: interrogate)} faire cela ? \\
\unkonly{} & {\color{red}\bf Puorquoi} {\color{orange}(typo)} faire cela ?\\ \hline\hline
Original& Si seulement je pouvais me muscler {\bf aussi} rapidement.\\
English gloss& If only I could build my muscle {\bf this} fast.\\
\unconstrained{} & Si seulement je pouvais me muscler {\color{red}\bf etc}  rapidement.\\
\knn{} & Si seulement je pouvais me muscler {\color{red}\bf plsu} {\color{orange}(typo for ``more'')} rapidement.\\
\unkonly{} & Si seulement je pouvais me muscler {\color{red}\bf asusi} {\color{orange}(typo)} rapidement.\\\hline\hline
\end{tabular}
}
\caption{\label{tab:qual_constraints} Examples of different adversarial inputs. The substituted word is highlighted.}
\end{table*}

We perform gradient-based attacks that replace one word in the sentence so as to maximize an adversarial loss function $\Ladv$, similar to the substitution attacks proposed in \cite{ebrahimi2018hotflip}.

\subsubsection{General Approach}

Precisely, for a word-based translation model $M$%
\footnote{Note that this formulation is also valid for character-based models (see \citet{Ebrahimi2018OnAE}) and subword-based models. For subword-based models, additional difficulty would be introduced due to changes to the input resulting in different subword segmentations. This poses an interesting challenge that is beyond the scope of the current work.}, and given an input sentence $w_1,\ldots,w_n$, we find the position $i^*$ and word $w^*$ satisfying the following optimization problem:

\begin{equation}\label{eq:adv_optim}
    \argmax_{1\leq i\leq n, \hat w\in \mathcal V}\Ladv(w_0,\ldots,w_{i-1},\hat w, w_{i+1},\ldots,w_n)
\end{equation}

\noindent where $\Ladv$ is a differentiable function which represents our adversarial objective. Using the first order approximation of $\Ladv$ around the original word vectors $\w_1,\ldots,\w_n$\footnote{More generally we will use the bold $\w$ when talking about the embedding vector of word $w$}, this can be derived to be equivalent to optimizing 

\begin{equation}
   \argmax_{1\leq i\leq n, \hat w\in \mathcal V}\left[\hat\w-{\w}_i\right]^\intercal\nabla_{\w_i}\Ladv 
\end{equation}

The above optimization problem can be solved by brute-force in $\bigO{n\vert\mathcal V\vert}$ space complexity, whereas the time complexity is bottlenecked by a $\vert\mathcal V\vert\times d$ times $n\times d$ matrix multiplication, which is not more computationally expensive than computing logits during the forward pass of the model. Overall, this naive approach is sufficiently fast to be conducive to adversarial training.
We also found that the attacks benefited from normalizing the gradient by taking its sign.

Extending this approach to finding the optimal perturbations for more than 1 substitution would require exhaustively searching over all possible combinations.
However, previous work \cite{Ebrahimi2018OnAE} suggests that greedy search is a good enough approximation.

\subsubsection{The Adversarial Loss $\Ladv$}

We want to find an adversarial input $\hat x$ such that, assuming that the model has produced the correct output $y_1,\ldots,y_{t-1}$ up to step $t-1$ during decoding, the probability that the model makes an error at the next step $t$ is maximized. 

In the log-semiring, this translates into the following loss function:
\begin{equation}
    \Ladv(\hat x, y)=\sum_{t=1}^{\vert y\vert}\log(1-p(y_t\mid \hat x, y_1,\ldots,y_{t-1}))
\end{equation}

\subsection{Enforcing Semantically Similar Adversarial Inputs}
\label{sec:constraints}

In contrast to previous methods, which don't consider meaning preservation, we
propose simple modifications of the approach presented in \S\ref{sec:attack_paradigm} to create adversarial perturbations at the word level that are more likely to preserve meaning.
The basic idea is to restrict the possible word substitutions to similar words. We compare two sets of constraints:

\textbf{\knn:} This constraint enforces that the word be replaced only with one of its 10 nearest neighbors in the source embedding space. This has two effects: first, the replacement will be likely semantically related to the original word (if words close in the embedding space are indeed semantically related, as hinted by Table~\ref{tab:qual_constraints}). Second, it ensures that the replacement's word vector is close enough to the original word vector that the first order assumption is more likely to be satisfied.

\textbf{\unkonly:} This constraint requires that the substituted words must be obtained by swapping characters. Word internal character swaps have been shown to not affect human readers greatly \cite{mccusker1981word}, hence making them likely to be meaning-preserving. Moreover we add the additional constraint that the substitution must not be in the vocabulary, which will likely be particularly meaning-destroying on the target side for the word-based models we test here. In such cases where word-internal character swaps are not possible or can't produce \ac{oov} words, we resort to the naive strategy of repeating the last character of the word. The exact procedure used to produce this kind of perturbations is described in Appendix \ref{sec:gen_char_swap}. Note that for a word-based model, every \ac{oov} will look the same (a special \unk{} token), however the choice of \ac{oov} will still have an influence on the output of the model because we use unk-replacement.

In contrast, we refer the base attack without constraints as {\bf \unconstrained} hereforth. Table \ref{tab:qual_constraints} gives qualitative examples of the kind of perturbations generated under the different constraints.

For subword-based models, we apply the same procedures at the subword-level  on the original segmentation. We then de-segment and re-segment the resulting sentence (because changes at the subword or character levels are likely to change the segmentation of the resulting sentence).

\section{Experiments}
\label{sec:experiments}


Our experiments serve two purposes.
First, we examine our proposed framework of evaluating adversarial attacks (\S\ref{sec:eval_adv_attacks}), and also elucidate which automatic metrics correlate better with human judgment for the purpose of evaluating adversarial attacks (\S\ref{sec:corr_human_auto}). Second, we use this evaluation framework to compare various adversarial attacks and demonstrate that adversarial attacks that are explicitly constrained to preserve meaning receive better assessment scores (\S\ref{sec:attack_results}).

\subsection{Experimental setting}

\textbf{Data:}
Following previous work on adversarial examples for \ac{seq2seq} models \cite{belinkov2018synthetic,Ebrahimi2018OnAE}, we perform all experiments on the IWSLT2016 dataset \cite{Cettolo2016TheI2} in the \{French,German,Czech\}$\rightarrow$English directions (\fren{}, \deen{} and \csen{}). We compile all previous IWSLT test sets before 2015 as validation data, and keep the 2015 and 2016 test sets as test data. The data is tokenized with the Moses tokenizer \cite{Koehn:2007:MOS:1557769.1557821}. The exact data statistics can be found in Appendix \ref{sec:iwslt2016_stats}.

\textbf{\ac{mt} Models:}
We perform experiments with two common \ac{nmt} models. The first is an LSTM based encoder-decoder architecture with attention~\citep{luong-pham-manning:2015:EMNLP}. It uses 2-layer encoders and decoders, and dot-product attention. We set the word embedding dimension to 300 and all others to 500.
The second model is a self-attentional Transformer \cite{vaswani2017attention}, with 6 1024-dimensional encoder and decoder layers and 512 dimensional word embeddings. Both the models are trained with Adam \cite{Kingma2014Adam}, dropout \cite{srivastava2014dropout} of probability 0.3 and label smoothing \cite{Szegedy2016RethinkingTI} with value 0.1. We experiment with both word based models (vocabulary size fixed at 40k) and subword based models (BPE \cite{sennrich-haddow-birch:2016:P16-12} with 30k operations). For word-based models, we perform \unk{} replacement, replacing \unk{} tokens in the translated sentences with the source words with the highest attention value during inference. The full experimental setup and source code are available at \codeurl{}.

\textbf{Automatic Metric Implementations:} To evaluate both sentence and corpus level BLEU score, we first de-tokenize the output and use {\tt sacreBLEU}\footnote{\url{https://github.com/mjpost/sacreBLEU}} \cite{post2018call} with its internal {\tt intl} tokenization, to keep BLEU scores agnostic to tokenization. We compute METEOR using the official implementation\footnote{\url{http://www.cs.cmu.edu/~alavie/METEOR/}}. ChrF is reported with the {\tt sacreBLEU} implementation on detokenized text with default parameters. A toolkit implementing the evaluation framework described in \S\ref{sec:adv_tradeoff} for these metrics is released at \teapoturl{}.

\begin{table*}[ht]
\centering
\begin{tabular}{clccccccc}
 &  & \multicolumn{3}{c}{LSTM} &\ & \multicolumn{3}{c}{Transformer} \\\hline\hline
 & Language pair & \csen & \deen & \fren&\ & \csen & \deen & \fren\\\cline{3-9}
\multirow{8}{*}{Word-based}& & \multicolumn{3}{c}{Target RDChrF} & & \multicolumn{3}{c}{Target RDChrF}\\\cline{3-5}\cline{7-9}
 & Original chrF & 45.68 & 49.43 & 57.49 & & 47.66 & 51.08 & 58.04\\
 &\unconstrained{} & 25.38 & 25.54 & 25.59 & & 25.24 & 25.00 & 24.68\\
 &\unkonly{} & 24.11 & 24.94 & 23.60 & & 21.59 & 23.23 & 21.75\\
 &\knn{} & 15.00 & 15.59 & 15.22& & 20.74 & 19.97 & 18.59 \\
 &&\multicolumn{3}{c}{Source \goodscore{}}&&\multicolumn{3}{c}{Source \goodscore{}}\\\cline{3-5}\cline{7-9}
 &\unconstrained{} & 70.14 & 72.39 & 74.29 & & 69.03 & 71.93 & 73.23\\
 &\unkonly{} & 82.65 & 84.40 & 86.62 & & 84.13 & 85.97 & 87.02\\
 &\knn{} & 78.08 & 78.11 & 77.62 & & 74.94 & 77.92 & 77.88\\\hline
\multirow{8}{*}{Subword-based}& & \multicolumn{3}{c}{Target RDChrF} & & \multicolumn{3}{c}{Target RDChrF}\\\cline{3-5}\cline{7-9}
 & Original chrF &48.30 & 52.42 & 59.08 & & 49.70 & 54.01 & 59.65\\
 &\unconstrained{} & 25.79 & 26.03 & 26.96 & & 23.97 & 25.07 & 25.28\\
 &\unkonly{} & 18.65 & 19.15 & 19.75 & & 16.98 & 18.38 & 17.85\\
 &\knn{} & 15.00 & 16.26 & 17.12 & & 19.02 & 18.58 & 18.63\\
 &&\multicolumn{3}{c}{Source \goodscore{}}&&\multicolumn{3}{c}{Source \goodscore{}}\\\cline{3-5}\cline{7-9}
 &\unconstrained{} & 69.32 & 72.12 & 73.57 & & 68.66 & 71.51 & 72.65\\
 &\unkonly{} & 85.84 & 87.46 & 87.98 & & 85.79 & 87.07 & 87.99\\
 &\knn{} & 76.17 & 77.74 & 78.03 & & 73.05 & 75.91 & 76.54\\
\hline
\end{tabular}
\caption{\label{tab:all_attacks_results} Target RDchrF and source \goodscore{} scores for all the attacks on all our models (word- and subword-based LSTM and Transformer).}
\end{table*}

\subsection{Correlation of Automatic Metrics with Human Judgment}
\label{sec:corr_human_auto}

We first examine which of the automatic metrics listed in \S\ref{sec:eval_metrics} correlates most with human judgment for our adversarial attacks. For this experiment, we restrict the scope to the case of the LSTM model on \fren{}. For the French side, we randomly select 900 sentence pairs $(x,\hat x)$ from the validation set, 300 for each of the \unconstrained{}, \knn{} and \unkonly{} constraints. To vary the level of perturbation, the 300 pairs contain an equal amount of perturbed input obtained by substituting 1, 2 and 3 words.
On the English side, we select 900 pairs of reference translations and translations of adversarial input $(y, \hat y_M)$ with the same distribution of attacks as the source side, as well as 300 $(y, y_M)$ pairs (to include translations from original inputs). This amounts to 1,200 sentence pairs in the target side.

These sentences are sent to English and French speaking annotators to be rated according to the guidelines described in \S\ref{sec:human_judgement}. Each sample (a pair of sentences) is rated by two independent evaluators. If the two ratings differ, the sample is sent to a third rater (an auditor and subject matter expert) who makes the final decision.

\begin{table}[!h]
\begin{tabular}{lccc}
Language & BLEU & METEOR & chrF\\\hline\hline
French&0.415&0.440&\bf0.586$^*$\\
English&0.357&0.478$^*$&\bf0.497\\\hline
\end{tabular}
\caption{\label{tab:human_eval_results} Correlation of automatic metrics to human judgment of adversarial source and target sentences. ``$^*$'' indicates that the correlation is significantly better than the next-best one.}
\end{table}

Finally, we compare the human results to each automatic metric with Pearson's correlation coefficient. The correlations are reported in Table \ref{tab:human_eval_results}. As evidenced by the results, \goodscore{} exhibits higher correlation with human judgment, followed by METEOR and BLEU. This is true both on the source side ($x$ vs $\hat x$) and in the target side ($y$ vs $\hat y_M$). We evaluate the statistical significance of this result using a paired bootstrap test for $p<0.01$. Notably we find that chrF is significantly better than METEOR in French but not in English. This is not too unexpected because METEOR has access to more language-dependent resources in English (specifically synonym information) and thereby can make more informed matches of these synonymous words and phrases. Moreover the French source side contains more ``character-level'' errors (from \unkonly{} attacks) which are not picked-up well by word-based metrics like BLEU and METEOR. For a breakdown of the correlation coefficients according to number of perturbation and type of constraints, we refer to Appendix \ref{sec:human_breakdown}.

Thus, in the following, we report attack results both in terms of \goodscore{} in the source ($\ssrc$) and \ac{rdb} in the target ($d_{\text{tgt}}$).

\begin{figure*}[!t]
\centering
\includegraphics[width=0.8\textwidth]{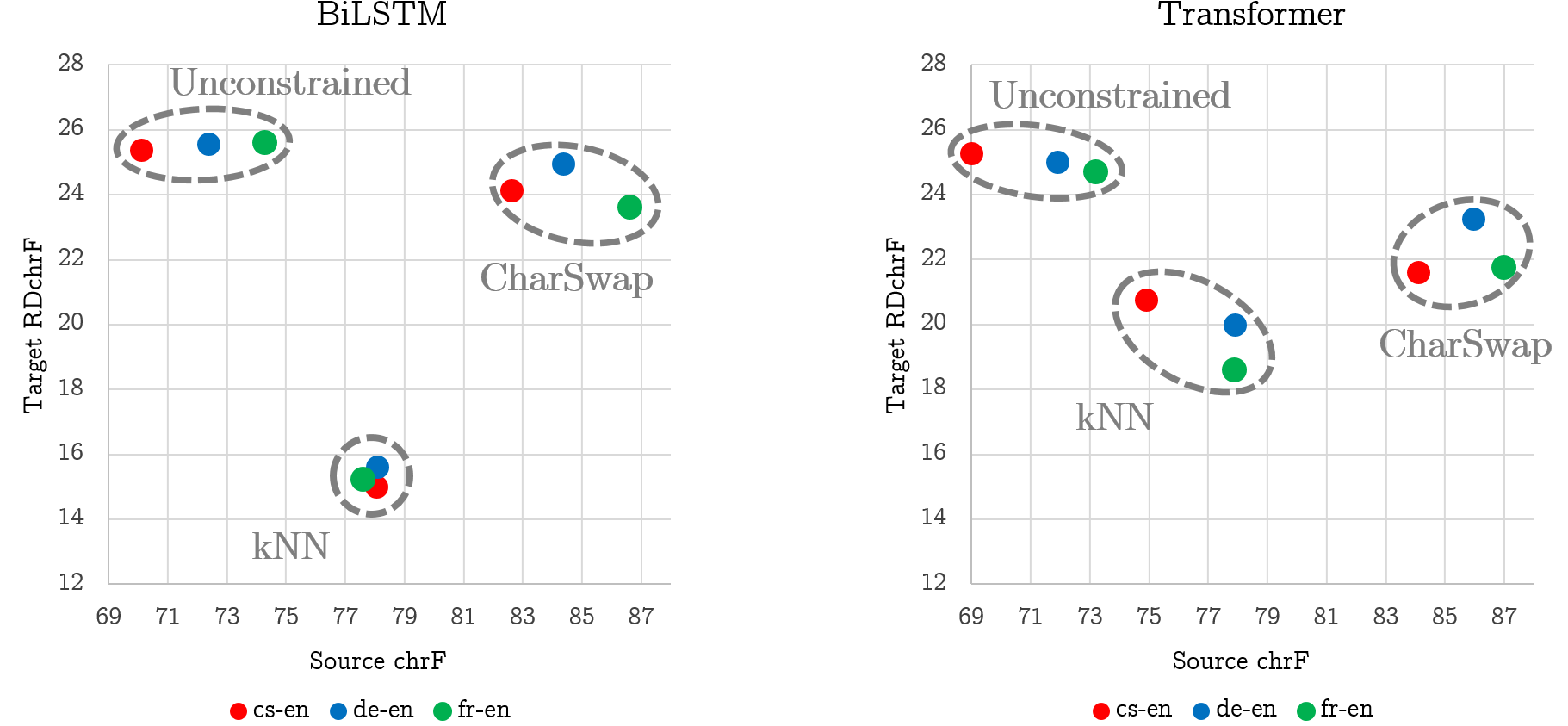}
\caption{\label{fig:chrf_plots} Graphical representation of the results in Table \ref{tab:all_attacks_results} for word-based models. High source \goodscore{} and target \ac{rdb} (upper-right corner) indicates a good attack.}
\end{figure*}

\subsection{Attack Results}
\label{sec:attack_results}

We can now compare attacks under the three constraints \unconstrained{}, \knn{} and \unkonly{} and draw conclusions on their capacity to preserve meaning in the source and destroy it in the target. Attacks are conducted on the validation set using the approach described in \S\ref{sec:attack_paradigm} with 3 substitutions (this means that each adversarial input is at edit distance at most 3 from the original input). Results (on a scale of 0 to 100 for readability) are reported in Table \ref{tab:all_attacks_results} for both word- and subword- based LSTM and Transformer models. To give a better idea of how the different variables (language pair, model, attack) affect performance, we give a graphical representation of these same results in Figure \ref{fig:chrf_plots} for the word-based models.
The rest of this section discusses the implication of these results.

\textbf{Source \goodscore{} Highlights the Effect of Adding Constraints:}
Comparing the \knn{} and \unkonly{} rows to \unconstrained{} in the ``source'' sections of Table \ref{tab:all_attacks_results} clearly shows that constrained attacks have a positive effect on meaning preservation. Beyond validating our assumptions from \S\ref{sec:constraints}, this shows that source \goodscore{} is useful to carry out the comparison in the first place\footnote{It can be argued that using \goodscore{} gives an advantage to \unkonly{} over \knn{} for source preservation (as opposed to METEOR for example). We find that this is the case for Czech and German (source METEOR is higher for \knn{}) but not French. Moreover we find (see \ref{sec:human_breakdown}) that \goodscore{} correlates better with human judgement even for \knn{}.}. To give a point of reference, results from the manual evaluation carried out in \S\ref{sec:corr_human_auto} show that that $90\%$ of the French sentence pairs to which humans gave a score of 4 or 5 in semantic similarity have a \goodscore{} $>78$.

\begin{table}[!h]
{
\small
\setlength\tabcolsep{2pt}
\begin{tabular}{ll}
\hline\hline
\multicolumn{2}{c}{Successful attack}\\
\multicolumn{2}{c}{(source \goodscore{} $=80.89$, target \ac{rdb} $=84.06$)}\\\hline
Original & Ils le r\'{e}investissent directement en engageant\\
& plus de proc\`{e}s.\\
Adv. src& {\color{red}Ilss} le r\'{e}investissent {\color{red}dierctement} en {\color{red}engagaent}\\
&plus de proc\`{e}s.\\
Ref. & They plow it right back into filing more troll \\
&lawsuits.\\
Base output& They direct it directly by engaging more cases.\\
Adv. output& .. de plus.\\\hline\hline
\multicolumn{2}{c}{Unsuccessful attack}\\
\multicolumn{2}{c}{(source \goodscore{} $=54.46$, target \ac{rdb} $=0.00$)}\\\hline
Original & C'\'{e}tait en Juillet 1969.\\
Adv. src & C'{\color{red}\'{e}tiat} en {\color{red}Jiullet} 1969.\\
Ref. & This is from July, 1969.\\
Base output & This was in July 1969.\\
Adv. output & This is. in 1969.\\\hline\hline
\end{tabular}
}
\caption{\label{tab:qual_unkonly_attack} Example of \unkonly{} attacks on the \fren{} LSTM. The first example is a successful attack (high source \goodscore{} and target \ac{rdb}) whereas the second is not.}
\end{table}

\textbf{Different Architectures are not Equal in the Face of Adversity:}
Inspection of the target-side results yields several interesting observations. First, the high \ac{rdb} of \unkonly{} for word-based model is yet another indication of their known shortcomings when presented with words out of their training vocabulary, even with \unk{}-replacement.
Second, and perhaps more interestingly, Transformer models appear to be less robust to small embedding perturbations (\knn{} attacks) compared to LSTMs. Although the exploration of the exact reasons for this phenomenon is beyond the scope of this work, this is a good example that \ac{rdb} can shed light on the different behavior of different architectures when confronted with adversarial input.
Overall, we find that the \unkonly{} constraint is the only one that consistently produces attacks with $>1$ average success (as defined in Section \ref{sec:adv_tradeoff}) according to Table \ref{tab:all_attacks_results}. Table \ref{tab:qual_unkonly_attack} contains two qualitative examples of this attack on the LSTM  model in \fren{}.

\section{Adversarial Training with Meaning-Preserving Attacks}
\label{sec:adv_train}

\subsection{Adversarial Training}

Adversarial training \cite{Goodfellow2014ExplainingAH} augments the training data with adversarial examples. Formally, in place of the \ac{nll} objective on a sample $x, y$, $\mathcal{L}(x,y)=NLL(x,y)$, the loss function is replaced with an interpolation of the \ac{nll} of the original sample $x,y$ and an adversarial sample $\hat x, y$:

\begin{equation}
    \mathcal{L}'(x,y)=(1-\alpha)NLL(x,y) + \alpha NLL(\hat x,y)
\end{equation}

\citet{Ebrahimi2018OnAE} suggest that while adversarial training improves robustness to adversarial attacks, it can be detrimental to test performance on non-adversarial input.
We investigate whether this is still the case when adversarial attacks are largely meaning-preserving.

In our experiments, we generate $\hat x$ by applying 3 perturbations on the fly at each training step. To maintain training speed we do not solve Equation (\ref{eq:adv_optim}) iteratively but in one shot by replacing the argmax by top-3. Although this is less exact than iterating, this makes adversarial training time less than $2\times$ slower than normal training. We perform adversarial training with perturbations without constraints (\unconstrained{}-adv) and with the \unkonly{} constraint (\unkonly{}-adv). All experiments are conducted with the word-based LSTM model.

\subsection{Results}
\label{sec:adv_train_experiments}

Test performance on non-adversarial input is reported in Table \ref{tab:adv_train_bleu_scores}. In keeping with the rest of the paper, we primarily report \goodscore{} results, but also show the standard BLEU as well.

We observe that when $\alpha=1.0$, \ie{}\ the model only sees the perturbed input during training\footnote{This setting is reminiscent of word dropout \cite{iyyer-EtAl:2015:ACL-IJCNLP}.}, the \unconstrained{}-adv model suffers a drop in test performance, whereas \unkonly{}-adv's performance is on par with the original. This is likely attributable to the spurious training samples $(\hat x, y)$ where $y$ is not an acceptable translation of $\hat x$ introduced by the lack of constraint. This effect disappears when $\alpha=0.5$ because the model sees the original samples as well.

Not unexpectedly, Table \ref{tab:adv_train_robustness} indicates that \unkonly{}-adv is more robust to \unkonly{} constrained attacks for both values of $\alpha$, with $1.0$ giving the best results. On the other hand, \unconstrained{}-adv is similarly or more vulnerable to these attacks than the baseline.
Hence, we can safely conclude that adversarial training with \unkonly{} attacks improves robustness while not impacting test performance as much as unconstrained attacks.

\begin{table}[t]
\centering
\begin{tabular}{lccc}
Language pair & \csen & \deen & \fren\\\hline\hline
\multirow{2}{*}{Base} & 44.21 & 49.30 & 55.67  \\
& \small (22.89) & \small (28.61) & \small (35.28)\\\cline{2-4}
 & \multicolumn{3}{c}{$\alpha=1.0$} \\\cline{2-4}
\multirow{2}{*}{\unconstrained{}{}-adv} & 41.38 & 46.15 & 53.39 \\
& \small (21.51) & \small (27.06) & \small (33.96)\\
\multirow{2}{*}{\unkonly{}-adv} & 43.74 & 48.85 & 55.60 \\
& \small (23.00) & \small (28.45) & \small (35.33)\\\cline{2-4}
 & \multicolumn{3}{c}{$\alpha=0.5$} \\\cline{2-4}
\multirow{2}{*}{\unconstrained{}{}-adv} &  43.68 & 48.60 & 55.55 \\
& \small (22.93) & \small (28.30) & \small (35.25)\\
\multirow{2}{*}{\unkonly{}-adv} & 44.57 & 49.14 & 55.88 \\
& \small (23.66) & \small (28.66) & \small (35.63)\\
\hline
\end{tabular}
\caption{\label{tab:adv_train_bleu_scores}\goodscore{} (BLEU) scores on the original test set before/after adversarial training of the word-based LSTM model.}
\end{table}

\begin{table}[t]
\centering
\begin{tabular}{lccc}
Language pair & \csen & \deen & \fren\\\hline\hline
Base &  24.11 & 24.94 & 23.60   \\\cline{2-4}
 & \multicolumn{3}{c}{$\alpha=1.0$} \\\cline{2-4}
\unconstrained{}-adv &  25.99 & 26.24 & 25.67  \\
\unkonly{}-adv &  16.46 & 17.19 & 15.72  \\\cline{2-4}
 & \multicolumn{3}{c}{$\alpha=0.5$} \\\cline{2-4}
\unconstrained{}-adv & 26.52 & 27.26 & 24.92  \\
\unkonly{}-adv & 20.41 & 20.24 & 16.08 \\
\hline
\end{tabular}
\caption{\label{tab:adv_train_robustness} Robustness to \unkonly{} attacks on the validation set with/without adversarial training (\ac{rdb}). Lower is better.}
\end{table}
\section{Related work}
\label{sec:related}

Following seminal work on adversarial attacks by \citet{Szegedy2013IntriguingPO}, 
\citet{Goodfellow2014ExplainingAH} introduced gradient-based attacks and adversarial training. Since then, a variety of attack \cite{MoosaviDezfooli2016DeepFoolAS} and defense \cite{Ciss2017ParsevalNI,Kolter2017ProvableDA} mechanisms have been proposed.
Adversarial examples for \ac{nlp} specifically have seen attacks on sentiment \cite{papernot2016crafting,samanta2017towards,ebrahimi2018hotflip}, malware \cite{grosse2016adversarial}, gender \cite{reddy-knight:2016:NLPandCSS} or toxicity \cite{hosseini2017deceiving} classification to cite a few.

In \ac{mt}, methods have been proposed to attack word-based \cite{zhao2018generating,cheng2018seq2sick} and character-based \cite{belinkov2018synthetic,Ebrahimi2018OnAE} models. However these works side-step the question of meaning preservation in the source: they mostly focus on target side evaluation. Finally there is work centered around meaning-preserving adversarial attacks for \ac{nlp} via paraphrase generation \cite{iyyer-EtAl:2018:N18-1} or rule-based approaches \cite{jia-liang:2017:EMNLP2017,ribeiro-singh-guestrin:2018:Long,naik-EtAl:2018:C18-1,alzantot-EtAl:2018:EMNLP}. However the proposed attacks are highly engineered and focused on English.

\section{Conclusion}

This paper highlights the importance of performing \emph{meaning-preserving} adversarial perturbations for \ac{nlp} models (with a focus on \ac{seq2seq}).
We proposed a general evaluation framework for adversarial perturbations and compared various automatic metrics as proxies for human judgment to instantiate this framework.
We then confirmed that, in the context of \ac{mt}, ``naive'' attacks do not preserve meaning in general, and proposed alternatives to remedy this issue.
Finally, we have shown the utility of adversarial training in this paradigm. 
We hope that this helps future work in this area of research to evaluate meaning conservation more consistently.

\section*{Acknowledgments}

The authors would like to extend their thanks to members of the LATTE team at Facebook and Neulab at Carnegie Mellon University for valuable discussions, as well as the anonymous reviewers for their insightful feedback. This research was partially funded by Facebook.

\bibliography{0_main}
\bibliographystyle{acl_natbib}

\clearpage
\appendix

\section{Supplemental Material}
\label{sec:supplemental}
\subsection{Generating \ac{oov} Replacements with Internal Character Swaps}
\label{sec:gen_char_swap}

We use the following snippet to produce an \ac{oov} word from an existing word:

\begin{lstlisting}[language=Python]
def make_oov(
  word,
  vocab,
  max_scrambling,
):
  """Modify a word to make it OOV
  (while keeping the meaning)"""
  # If the word has >3 letters
  # try scrambling them
  L = len(word)
  if L > 3:
    # For a fixed number of steps
    for _ in range(max_scrambling):
      # Swap two adjacent letters
      # in the middle of the word
      pos = random.randint(1, L - 3)
      word = word[:pos] 
      word += word[pos+1] + word[pos]
      word += word[pos+2:]
      # If we got an OOV already just
      # return it
      if word not in vocab:
        return word
  # If nothing worked, or the word is
  # too short for scrambling, just
  # repeat the last letter ad nauseam
  char = word[-1]
  while word in vocab:
    word = word + char
  return word
\end{lstlisting}

\subsection{IWSLT2016 Dataset}
\label{sec:iwslt2016_stats}

See table \ref{tab:iwslt2016_stats} for statistics on the size of the IWSLT2016 corpus used in our experiments.

\begin{table}[!h]
\centering
\begin{tabular}{lccc}
& \#train & \#valid & \#test \\ \hline
\fren & 220.4k & 6,824 & 2,213 \\
\deen & 196.9k & 11,825 & 2,213 \\
\csen & 114.4k & 5,716 & 2,213\\
\hline
\end{tabular}
\caption{\label{tab:iwslt2016_stats}IWSLT2016 data statistics.
}
\end{table}

\subsection{Breakdown of Correlation with Human Judgement}
\label{sec:human_breakdown}

We provide a breakdown of the correlation coefficients of automatic metrics with human judgment for source-side meaning-preservation, both in terms of number of perturbed words (Table \ref{tab:corr_edits_breakdown}) and constraint (Table \ref{tab:corr_constraints_breakdown}). While those coefficients are computed on a much smaller sample size, and their differences are not all statistically significant with $p<0.01$, they exhibit the same trend as the results from Table \ref{tab:human_eval_results} (BLEU $<$ METEOR $<$ chrF). In particular Table \ref{tab:corr_edits_breakdown} shows that the good correlation of chrF with human judgment is not only due to the ability to distinguish between different number of edits.

\begin{table}
\begin{tabular}{cccc}
\# edits & BLEU & METEOR & chrF\\\hline\hline
1&0.351&0.352&\bf0.486$^*$\\
2&0.403&0.424&\bf0.588$^*$\\
3&0.334&0.393&\bf0.560$^*$\\
\hline
\end{tabular}
\caption{\label{tab:corr_edits_breakdown} Correlation of automatic metrics to human judgment of semantic similarity between original and adversarial source sentences, broken down by number of perturbed words. ``$^*$'' indicates that the correlation is significantly better than the next-best one.}
\end{table}

\begin{table}
\begin{tabular}{cccc}
Constraint & BLEU & METEOR & chrF\\\hline\hline
\unconstrained{} &0.274 & 0.572 & \bf0.599\\
\unkonly{}&0.274 & 0.319 & \bf0.383\\
\knn{} & 0.534 & 0.584 & \bf0.606\\
\hline
\end{tabular}
\caption{\label{tab:corr_constraints_breakdown} Correlation of automatic metrics to human judgment of semantic similarity between original and adversarial source sentences, broken down by type of constraint on the perturbation. ``$^*$'' indicates that the correlation is significantly better than the next-best one.}
\end{table}

\end{document}